\documentclass{article}
\usepackage[T1]{fontenc}
\usepackage{preprint,times}

\usepackage{amsmath,amsfonts,bm}

\def\eqref#1{equation~\ref{#1}}

\def\1{\bm{1}}

\DeclareMathAlphabet{\mathsfit}{\encodingdefault}{\sfdefault}{m}{sl}
\SetMathAlphabet{\mathsfit}{bold}{\encodingdefault}{\sfdefault}{bx}{n}

\usepackage{hyperref}
\usepackage{url}
\usepackage{graphicx}
\usepackage{booktabs}
\usepackage{wrapfig}
\usepackage{xcolor}
\usepackage{colortbl}
\usepackage{multirow}
\usepackage{tcolorbox}
\tcbuselibrary{skins,breakable}
\newtcolorbox{promptbox}[1]{enhanced, breakable, colback=black!2, colframe=black!55, colbacktitle=black!10, coltitle=black, fonttitle=\footnotesize\bfseries, title={#1},
  boxrule=0.4pt, arc=1pt, left=4pt, right=4pt, top=3pt, bottom=3pt, fontupper=\scriptsize\ttfamily\raggedright\frenchspacing, fontlower=\scriptsize\ttfamily\raggedright\frenchspacing,
  segmentation style={black!40, dashed}, before skip=8pt, after skip=8pt}
\newcommand{\promptvar}[1]{{\normalfont\itshape$\langle$#1$\rangle$}}
\newcommand{\promptlabel}[1]{{\normalfont\itshape #1:}\par\smallskip}
\definecolor{ourscol}{RGB}{236,241,250}
\hypersetup{hidelinks}

\title{From Uncertainty to Action:\\Learning to Steer LLM Agents}
\author{Hanwen Li$^{1}$, Jinhao Duan$^{2}$, Guanhua Zhu$^{1}$, Junchi Lu$^{3}$, \\
\textbf{Bo Shen$^{1}$, Chenxi Yuan$^{1,*}$, Kaidi Xu$^{4}$} \\
{\normalfont $^{1}$New Jersey Institute of Technology \quad $^{2}$University of North Carolina at Chapel Hill} \\
{\normalfont $^{3}$University of California, Irvine \quad $^{4}$City University of Hong Kong}}

\begin{document}
\maketitle
{\renewcommand{\thefootnote}{}\footnotetext{$^{*}$Corresponding author: \texttt{cy324@njit.edu}}}

\begin{abstract}
Steering an LLM agent means deciding whether to correct it, at which step, and with which mechanism.
Uncertainty is often used to decide when to correct an agent, but whether it can guide these decisions remains unclear.
We steer agent trajectories separately at every non-terminal step with each of four mechanisms and run each continuation to completion.
The resulting stepwise outcome table (SOT) holds about $82{,}000$ counterfactual continuations of $1{,}864$ trajectories from three benchmarks and two agents.
It shows that uncertainty can identify failing trajectories, but that no single signal reliably locates the step at which steering helps.
We therefore propose VoS (Value of Steering), a trajectory-level monitor, offline or online, that learns from SOT the value of steering at each step and decides where to steer by it.
A harm-budgeted trigger decides whether to steer, limiting the fraction of successful trajectories that VoS disturbs.
VoS improves on unmodified execution in all $12$ settings of benchmark, agent, and offline or online use, by $7.8$ points on average, and outperforms the strongest of five existing uncertainty-triggered methods in $11$, by $2.9$ points on average.
Ablations show that training on measured outcomes and a tight harm budget are both essential.

\end{abstract}

\section{Introduction}
\label{sec:introduction}

Uncertainty quantification (UQ) for large language models (LLMs) has often centered on answer correctness, hallucination detection, and confidence calibration~\citep{kadavath2022language,lin2022teaching,kuhn2023semantic,manakul2023selfcheckgpt,tian2023calibration,xiong2024can,farquhar2024semantic,geng2024survey}, and is typically assessed through error detection, reliability ranking, or calibration metrics~\citep{hendrycks2017baseline,guo2017calibration,malinin2021uncertainty,vashurin2025benchmarking}.
As LLMs are increasingly deployed as agents that plan, use tools, and interact with environments to complete multi-step tasks~\citep{yao2023react,schick2023toolformer,wang2024codeact,wang2024survey}, uncertainty estimates should also help improve task execution, not only assess the reliability of model outputs.

Multi-step execution creates opportunities to correct an agent before a task fails, since early errors otherwise propagate through the remaining steps~\citep{cemri2025why,zhang2025which}, and prior work offers mechanisms for doing so, from prompting an agent to reflect on its own progress to supplying external feedback and assistance~\citep{shinn2023reflexion,madaan2023selfrefine,kim2025reflact,gou2024critic,chen2023frugalgpt,zhao2024expel,wang2024awm}.
We call a targeted modification of an agent's subsequent execution at a selected point of its trajectory \emph{steering}.
Steering at every step adds computation and may disrupt otherwise successful execution~\citep{huang2024cannot,kamoi2024when}, so it should be used selectively, and uncertainty is a natural signal for this selection, as in selective prediction and deferral to an expert~\citep{elyaniv2010foundations,geifman2017selective,mozannar2020consistent}.

This requires three decisions: \emph{whether} to steer a trajectory, \emph{where}, that is, at which step, and with \emph{which} mechanism.
These decisions depend on whether a particular correction would improve the outcome, which the predicted risk of failure alone cannot tell: a likely failure may not be recoverable, a correction to a successful trajectory can introduce new errors, and the most uncertain step need not offer the greatest benefit.
We therefore ask: \emph{Can uncertainty guide effective steering of LLM agents?}

\begin{figure}[t]
\centering
\includegraphics[width=0.94\linewidth]{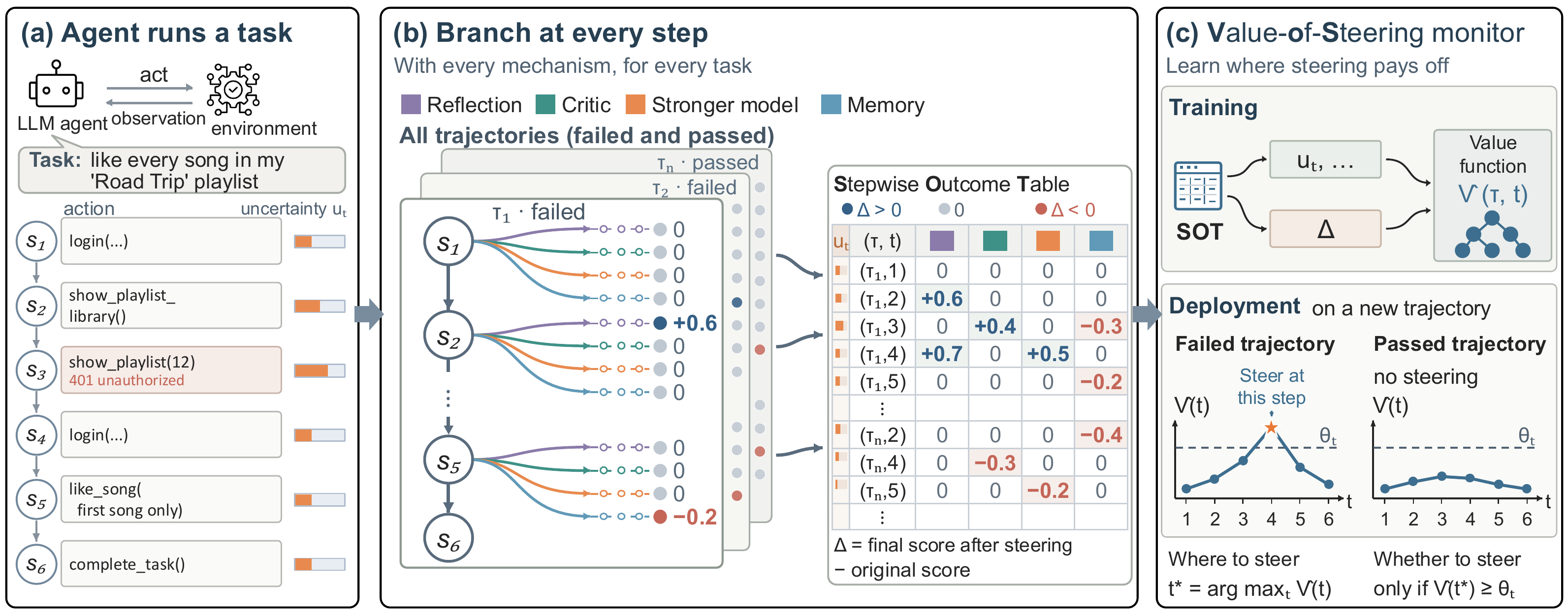}
\caption{Overview. (a)~An agent with step-level uncertainty. (b)~SOT: every step steered with every mechanism and run to the end. (c)~VoS: where to steer by a value function learned from SOT, whether by a threshold.}
\label{fig:overview}
\end{figure}

Among uncertainty-triggered methods for agents, Agentic UQ (AUQ)~\citep{zhang2026agentic} combines confidence-triggered reflection with memory, and ReDAct~\citep{piatrashyn2026redact} defers uncertain actions to a stronger model.
Their aggregate task gains, however, do not isolate the value of choosing the trajectory, the step, and the mechanism.
DIAL~\citep{li2026same} and \citet{zhang2026calibration} begin to address this with counterfactual comparisons, but only at the states that a trigger selects, whereas deciding where to steer requires comparing the outcomes of steering at different steps of the same trajectory.
We therefore introduce a stepwise counterfactual evaluation that steers every step of each trajectory with each of four mechanisms and records the change in the final task score in a stepwise outcome table (SOT), including the choices that a given trigger would not have made.
SOT captures both recovery on failed trajectories and harm to successful ones, and it lets us compare predictive metrics with steering gains as well as train and calibrate monitors.

Our results show that uncertainty helps identify failing trajectories but remains an imperfect guide to the value of correction.
That accurate failure prediction need not yield effective intervention has been noted for agent oversight in general~\citep{vasudev2026accurate,zhang2026calibration}; our stepwise outcomes show how this gap arises for uncertainty signals: a higher failure-detection AUROC does not yield greater task improvements, and the order in which a signal ranks the steps of a trajectory barely matches their gain.
The self-evaluation score $P(\mathrm{True})$~\citep{kadavath2022language}, for example, is among the weakest failure detectors in our study, yet it is the signal most consistently aligned with where steering helps.
The usefulness of a signal also depends on the mechanism: the four mechanisms agree moderately on which trajectories are recoverable but barely on which step to steer.

These findings shape our monitor, VoS (Value of Steering).
Since uncertainty carries a useful but imperfect failure signal and steering a successful trajectory can only lower its score, \emph{whether} to steer is decided by a harm-budgeted trigger, calibrated to limit the fraction of successful trajectories that are steered; since no single signal reliably locates the step at which steering helps, \emph{where} to steer is decided by a value function learned from all signals, with the measured outcome of steering as the target.
VoS runs offline, rolling back a completed execution before its result is accepted, or online, steering during execution; Figure~\ref{fig:overview} gives an overview.

Our contributions are:
\begin{itemize}
    \item We present SOT, a stepwise table of about $82{,}000$ counterfactual steering outcomes on $1{,}864$ trajectories, which measures harm to successful trajectories as well as recovery on failed ones. With it, we show that uncertainty helps identify failing trajectories, but that no single signal reliably locates the step at which steering helps or tells which mechanism to use.
    \item We propose VoS, a trajectory-level, harm-budgeted steering monitor that works during execution or by rolling back afterwards and decides where to steer by the value of steering at each step, estimated from uncertainty signals.
    \item On three benchmarks with two agents, offline and online, VoS improves on unmodified execution in all $12$ settings, by $7.8$ points on average, and outperforms the strongest of five uncertainty-triggered methods from prior work in $11$, by $2.9$ points on average. Ablations show that measured outcomes are a better training target than failure, binary, or process-reward labels, that a tight harm budget beats steering more trajectories, and that no single family of our $14$ estimators suffices across settings.
\end{itemize}

\section{Stepwise Counterfactual Evaluation of Steering}
\label{sec:evaluation}

To judge whether, where, and with which mechanism to steer, we measure what steering achieves at every step of every trajectory, not only at the steps that some trigger selects, and evaluate steering policies on the resulting outcomes.

\subsection{Setting and Notation}
\label{sec:setting}

An LLM agent solves a task over multiple steps~\citep{yao2023react,wang2024survey,ma2024agentboard,yao2024taubench}: at each step it reasons about the current state, issues an action such as a code block or a tool call~\citep{schick2023toolformer,wang2024codeact}, and receives an observation.
We call the unmodified execution of a task its \emph{reference trajectory}, $\tau = (s_1, \dots, s_N)$, where $s_t$ collects the reasoning, action, and observation of step $t$, and write $\tau_{\le t}$ for its first $t$ steps.
The benchmark's evaluator assigns it a task score $S(\tau) \in [0, 1]$ and marks it as successful or failed.
Each step also carries a vector of uncertainty scores $u_t \in \mathbb{R}^K$, computed from the token distributions of the agent's model when it produced step $t$ and oriented so that higher values indicate greater uncertainty.

\subsection{Steering Mechanisms}
\label{sec:mechanisms}

A steering mechanism $m$ acts at a single step $t$: given the task instruction and the prefix $\tau_{\le t}$, it writes one message, or abstains, and the message is inserted into the agent's context before step $t+1$, after which the agent continues unchanged.
The four mechanisms $\mathcal{M}$ that we study share this interface, so they can be compared at the same steps of the same trajectories.
\textbf{Reflection} follows ReflAct~\citep{kim2025reflact}, a goal-grounded form of self-reflection~\citep{shinn2023reflexion,madaan2023selfrefine}: the agent's own model reflects on the current state in relation to the task goal, and the message adds an instruction to reflect in the same way before every later action.
\textbf{Critic feedback} follows CRITIC~\citep{gou2024critic}: a critic verifies the trajectory so far in a copy of the environment restored to step $t$ and sends its critique with a corrected next action, or abstains if it finds nothing wrong.
\textbf{Stronger-model assistance} follows the deferred step of ReDAct~\citep{piatrashyn2026redact}: a stronger model writes the next action with its reasoning, and the agent executes this action as given.
\textbf{Memory-based guidance}, in the spirit of experience-based guidance~\citep{shinn2023reflexion,zhao2024expel,wang2024awm}, has an advisor read the prefix together with a note of recurring failure patterns, synthesized offline from failed training trajectories, and send a diagnosis and suggested next actions, followed by the note.
Only stronger-model assistance calls an external model; Appendix~\ref{app:mechanisms} gives the prompts.

\subsection{Measuring Steering Outcomes}
\label{sec:outcomes}

We write $\tau^{m,t}$ for the execution that results from steering $\tau$ at step $t$ with mechanism $m$, and define the \emph{steering outcome} as the change in task score,
\begin{equation}
    \Delta_m(\tau, t) = S\big(\tau^{m,t}\big) - S(\tau).
    \label{eq:outcome}
\end{equation}
On a successful reference trajectory, which already attains the full score, the outcome can only be zero or negative, so these trajectories measure the harm that steering can do.
To measure an outcome, we replay the recorded steps $1, \dots, t$, insert the message, and let the agent run to completion; if the mechanism abstains, the agent continues without a message.

We measure the outcome at every non-terminal step $t < N$ of every reference trajectory, failed or successful, and for all four mechanisms.
The resulting \emph{stepwise outcome table} (SOT) $\mathcal{D} = \{\Delta_m(\tau, t)\}$ thus also covers the choices that a particular triggering rule would not have made.
Each entry is observed by re-executing the agent from the same prefix, not inferred from a causal model or estimated from logged decisions~\citep{dudik2011doubly,bottou2013counterfactual}.

\subsection{Steering Policies and Evaluation}
\label{sec:policies}

In our main setting, a steering policy $\pi$ steers each trajectory at most once: it either leaves the trajectory alone, $\pi(\tau) = \emptyset$, or selects a step and a mechanism, $\pi(\tau) = (t, m)$ with $t < N$ and $m \in \mathcal{M}$.
It thus makes three decisions: \emph{whether} to steer a trajectory, \emph{where} to steer it, and \emph{which} mechanism to use.
An \emph{offline} policy sees the complete reference trajectory, but not its score, and rolls back to the step it selects; an \emph{online} policy decides at each step, from the prefix $\tau_{\le t}$ alone, whether to steer now.
A \emph{monitor} is a policy whose decisions depend on the trajectory only through the uncertainty scores and the step index: $u_1, \dots, u_N$ offline and $u_1, \dots, u_t$ online.

\par\textbf{Evaluation and cost.}\quad
We evaluate a policy by the task score it attains over all test trajectories, successful and failed, so that harm to successful trajectories counts against recovery on failed ones; the fraction of successful trajectories that it steers is its \emph{false-trigger rate}.
The cost $C(\pi)$ counts the additional agent steps executed because of steering, plus the mechanism's own call in agent-step equivalents of its tokens, averaged over the trajectories: online, only the difference to the replaced rest of the reference trajectory counts; offline, every step after the rollback.

\subsection{Benchmarks, Agents, and Uncertainty Estimators}
\label{sec:data}

\begin{figure}[t]
\centering
\begin{minipage}[b]{0.3290\linewidth}\centering
\includegraphics[width=\linewidth]{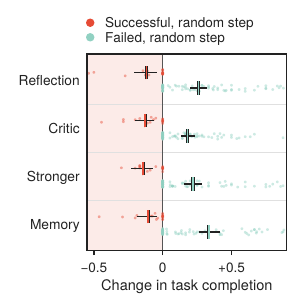}\\[1pt]{\small (a)}
\end{minipage}\hfill
\begin{minipage}[b]{0.2087\linewidth}\centering
\includegraphics[width=\linewidth]{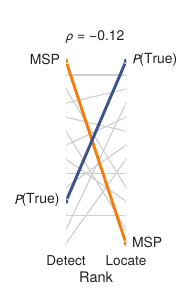}\\[1pt]{\small (b)}
\end{minipage}\hfill
\begin{minipage}[b]{0.3901\linewidth}\centering
\includegraphics[width=\linewidth]{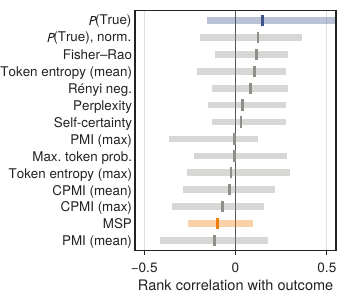}\\[1pt]{\small (c)}
\end{minipage}
\caption{AppWorld, Gemma; Figure~\ref{fig:s3_qwen} shows Qwen. (a)~Change in task completion per trajectory when steered at a random step; ticks: mean with 95\% bootstrap interval. Steering lowers the score of successful trajectories with every mechanism. (b)~Estimators ranked by failure detection and by within-trajectory localization; $\rho$: Spearman correlation. The best detectors are not the best locators. (c)~Per-trajectory Spearman correlation between each estimator and the outcome of memory-based guidance across steps; bar: interquartile range, tick: median. No estimator orders the steps like their outcomes.}
\label{fig:s3}
\end{figure}

\par\textbf{Benchmarks and agents.}\quad
Our main benchmark is AppWorld~\citep{trivedi2024appworld}, in which an agent writes code against the APIs of simulated apps and unit tests check their final state; the task score is one if all tests pass and zero otherwise, as in the task goal completion that we report, and VoS is trained on the fraction of tests passed.
In ALFWorld~\citep{shridhar2021alfworld}, an agent completes household tasks in a text environment and wins or loses; in WebShop~\citep{yao2022webshop}, it buys a product in a simulated store, scored by its match with the instruction, and succeeds at a score of one.
The agents are Gemma-4-31B~\citep{gemmateam2026gemma4} and Qwen3.6-35B~\citep{yang2025qwen3,qwen2026qwen36}, run as ReAct-style agents~\citep{yao2023react} with open weights, so that token-level uncertainty can be computed at every step.
Stronger-model assistance uses DeepSeek-V4-Flash~\citep{deepseekai2026v4}.

\par\textbf{Uncertainty estimators.}\quad
At every step we compute 14 estimators as defined in LM-Polygraph~\citep{fadeeva2023lmpolygraph,vashurin2025benchmarking} with minor adaptations: sequence- and token-level probabilities~\citep{fomicheva2020unsupervised,malinin2021uncertainty}, entropy-based and information-geometric measures of the token distributions~\citep{malinin2021uncertainty,darrin2023rainproof,kang2025selfcertainty}, pointwise mutual information~\citep{takayama2019relevant,vanderpoel2022mutual}, and self-evaluation with $P(\mathrm{True})$~\citep{kadavath2022language}; Appendix~\ref{app:monitor} lists them.
All come from the token distributions of the step that the agent produced, with at most a few additional scoring passes, whose cost Table~\ref{tab:uq_cost} reports; we exclude sampling-based estimators~\citep{kuhn2023semantic,farquhar2024semantic,manakul2023selfcheckgpt,lin2024generating}, which need several generations at every step, and density-based ones~\citep{ren2023ood}, which would be fitted and scored on the same data.

\section{What Uncertainty Tells a Steering Policy}
\label{sec:findings}

We now use SOT to ask what uncertainty contributes to each of the three decisions, using each signal on its own as a probe, on completed AppWorld trajectories.

\subsection{Whether: Uncertainty Identifies Failing Trajectories}
\label{sec:whether}

\textbf{Uncertainty carries a trajectory-level failure signal.}
A classifier on summary statistics of the 14 estimators over a completed trajectory detects failed trajectories with an AUROC of up to $0.73$ on Gemma and $0.71$ on Qwen, in line with earlier reports that uncertainty predicts the failure of agents~\citep{zhao2025saup,duan2025uprop,barbi2025preventing,zhang2026agentic}, so a monitor can use it to decide whether to steer.
The signal is imperfect, however, and steering is not safe by default: on successful trajectories, steering at a random step lowers the score with every mechanism, on average by up to $0.14$ on Gemma and $0.32$ on Qwen, as Figures~\ref{fig:s3}a and~\ref{fig:s3_qwen}a show, so a monitor must also control how often it steers successful trajectories.

\subsection{Where: The Same Signal Plays Different Roles}
\label{sec:where}

\par\textbf{Detecting failure and locating the step are different abilities.}\quad
Figures~\ref{fig:s3}b and~\ref{fig:s3_qwen}b rank the signals by two abilities: detection, how well the mean of a signal over a trajectory separates failed from successful trajectories, and localization, how well it ranks the helpful steps of a failed trajectory first, measured by a within-trajectory AUROC for each mechanism.
The two rankings are unrelated or even opposed, with a rank correlation of $-0.12$ on Gemma and $-0.41$ on Qwen.
$P(\mathrm{True})$~\citep{kadavath2022language} is among the weakest failure detectors on Gemma, and its detection direction is reversed on Qwen ($0.44$), yet it is the signal most consistently aligned with where steering helps, with a within-trajectory AUROC above $0.5$ for all four mechanisms on both agents.

\par\textbf{Uncertainty orders the steps differently from their gain.}\quad
To indicate where to steer, a signal should rank first the step at which steering helps most.
The target is the outcome of steering at a step, not the correctness of the step, which is what locating the erroneous step of a solution or of a failed execution asks for~\citep{lightman2024verify,wang2024mathshepherd,zhang2025which}.
We compare, within each failed trajectory, the order of the steps by a signal with their order by measured outcome, without training any model.
The two orders are nearly unrelated, as Figures~\ref{fig:s3}c and~\ref{fig:s3_qwen}c show: for memory-based guidance, the median rank correlation per trajectory is highest for $P(\mathrm{True})$, at only $+0.15$ on both agents, and for every estimator it is negative on at least a quarter of the trajectories.
The signal still carries some information: the step that $P(\mathrm{True})$ ranks first is the best step of the trajectory more often than a random step, in all eight combinations of agent and mechanism.
\textbf{The value of a signal therefore depends on the decision it serves}: a policy can neither choose a signal by how well it detects failures nor simply steer at the step that a signal ranks first.
This argues for learning from all signals jointly, with the steering outcome itself as the target.

\subsection{Which: Mechanisms Disagree on Where to Steer}
\label{sec:which}

\textbf{The four mechanisms agree moderately on which trajectories are recoverable, but barely on where to steer them.}
Across trajectories, their best outcomes correlate at $0.17$ to $0.66$; within a trajectory, their orders of the steps correlate at only $0.02$ to $0.19$, as Figure~\ref{fig:mechanism_matrix} in the appendix shows.
A signal can even relate to the outcome with opposite signs for different mechanisms, as token entropy does on Gemma for memory-based guidance and stronger-model assistance, a reversal of the kind that \citet{li2026same} report across environments and models for the value of test-time compute.

\section{Steering Monitors}
\label{sec:monitors}

\begin{table}[t]
\caption{Main results on each benchmark's own metric in \% (task goal completion on AppWorld, success rate on ALFWorld, score on WebShop), as mean $\pm$ standard deviation. Gray: references, not ranked; bold and underline: best and second-best policy in each row.}
\label{tab:main}
\centering
\footnotesize
\setlength{\tabcolsep}{2.6pt}\renewcommand{\arraystretch}{1.08}
\begin{tabular}{@{}l cc ccccc >{\columncolor{ourscol}}c@{}}\toprule
& \multicolumn{2}{c}{\textcolor{gray}{References}} & \multicolumn{6}{c}{Steering policies}\\
\cmidrule(lr){2-3}\cmidrule(lr){4-9}
& \textcolor{gray}{Unmod.} & \textcolor{gray}{Always} & AUQ & DIAL & Barbi et al. & DS-MCM & ReDAct & VoS (ours)\\\midrule
\multicolumn{9}{@{}l}{\emph{Offline}}\\[1pt]
\multirow{2}{*}{AppWorld} \, {\scriptsize Gemma} & \textcolor{gray}{69.0} & \textcolor{gray}{71.6}\,{\scriptsize\textcolor{gray}{$\pm$1.5}} & 68.1\,{\scriptsize\textcolor{gray}{$\pm$1.0}} & 68.9\,{\scriptsize\textcolor{gray}{$\pm$0.3}} & 69.7\,{\scriptsize\textcolor{gray}{$\pm$1.2}} & \underline{70.4}\,{\scriptsize\textcolor{gray}{$\pm$0.9}} & 69.3\,{\scriptsize\textcolor{gray}{$\pm$0.7}} & \textbf{73.4}\,{\scriptsize\textcolor{gray}{$\pm$1.9}} \\
\hphantom{AppWorld} \, {\scriptsize Qwen} & \textcolor{gray}{65.5} & \textcolor{gray}{55.6}\,{\scriptsize\textcolor{gray}{$\pm$1.6}} & 65.1\,{\scriptsize\textcolor{gray}{$\pm$0.9}} & 65.4\,{\scriptsize\textcolor{gray}{$\pm$0.4}} & 65.3\,{\scriptsize\textcolor{gray}{$\pm$0.7}} & 64.8\,{\scriptsize\textcolor{gray}{$\pm$0.6}} & \underline{66.0}\,{\scriptsize\textcolor{gray}{$\pm$0.6}} & \textbf{66.5}\,{\scriptsize\textcolor{gray}{$\pm$0.8}} \\
\cmidrule(lr){1-1}\cmidrule(lr){2-3}\cmidrule(lr){4-9}
\multirow{2}{*}{ALFWorld} \, {\scriptsize Gemma} & \textcolor{gray}{74.5} & \textcolor{gray}{76.6}\,{\scriptsize\textcolor{gray}{$\pm$2.8}} & 74.7\,{\scriptsize\textcolor{gray}{$\pm$1.2}} & \underline{80.1}\,{\scriptsize\textcolor{gray}{$\pm$1.8}} & 80.0\,{\scriptsize\textcolor{gray}{$\pm$1.8}} & 78.4\,{\scriptsize\textcolor{gray}{$\pm$1.7}} & 76.1\,{\scriptsize\textcolor{gray}{$\pm$1.0}} & \textbf{85.2}\,{\scriptsize\textcolor{gray}{$\pm$1.6}} \\
\hphantom{ALFWorld} \, {\scriptsize Qwen} & \textcolor{gray}{52.5} & \textcolor{gray}{69.0}\,{\scriptsize\textcolor{gray}{$\pm$2.8}} & 62.4\,{\scriptsize\textcolor{gray}{$\pm$2.5}} & 62.0\,{\scriptsize\textcolor{gray}{$\pm$3.5}} & 61.0\,{\scriptsize\textcolor{gray}{$\pm$2.4}} & 57.0\,{\scriptsize\textcolor{gray}{$\pm$1.6}} & \underline{64.9}\,{\scriptsize\textcolor{gray}{$\pm$2.7}} & \textbf{78.4}\,{\scriptsize\textcolor{gray}{$\pm$3.2}} \\
\cmidrule(lr){1-1}\cmidrule(lr){2-3}\cmidrule(lr){4-9}
\multirow{2}{*}{WebShop} \, {\scriptsize Gemma} & \textcolor{gray}{66.7} & \textcolor{gray}{68.7}\,{\scriptsize\textcolor{gray}{$\pm$0.9}} & 67.1\,{\scriptsize\textcolor{gray}{$\pm$0.3}} & 68.3\,{\scriptsize\textcolor{gray}{$\pm$0.5}} & \underline{69.4}\,{\scriptsize\textcolor{gray}{$\pm$0.2}} & 67.6\,{\scriptsize\textcolor{gray}{$\pm$0.3}} & 68.3\,{\scriptsize\textcolor{gray}{$\pm$0.7}} & \textbf{72.3}\,{\scriptsize\textcolor{gray}{$\pm$1.1}} \\
\hphantom{WebShop} \, {\scriptsize Qwen} & \textcolor{gray}{62.5} & \textcolor{gray}{63.9}\,{\scriptsize\textcolor{gray}{$\pm$0.7}} & 61.2\,{\scriptsize\textcolor{gray}{$\pm$0.2}} & 63.3\,{\scriptsize\textcolor{gray}{$\pm$0.6}} & 63.6\,{\scriptsize\textcolor{gray}{$\pm$0.4}} & \underline{65.0}\,{\scriptsize\textcolor{gray}{$\pm$0.0}} & 63.2\,{\scriptsize\textcolor{gray}{$\pm$0.0}} & \textbf{67.1}\,{\scriptsize\textcolor{gray}{$\pm$0.9}} \\
\midrule
\multicolumn{9}{@{}l}{\emph{Online}}\\[1pt]
\multirow{2}{*}{AppWorld} \, {\scriptsize Gemma} & \textcolor{gray}{69.0} & \textcolor{gray}{71.6}\,{\scriptsize\textcolor{gray}{$\pm$1.5}} & 69.2\,{\scriptsize\textcolor{gray}{$\pm$1.0}} & 70.0\,{\scriptsize\textcolor{gray}{$\pm$1.7}} & 69.7\,{\scriptsize\textcolor{gray}{$\pm$0.8}} & \underline{71.2}\,{\scriptsize\textcolor{gray}{$\pm$0.9}} & 69.4\,{\scriptsize\textcolor{gray}{$\pm$1.1}} & \textbf{73.6}\,{\scriptsize\textcolor{gray}{$\pm$1.8}} \\
\hphantom{AppWorld} \, {\scriptsize Qwen} & \textcolor{gray}{65.5} & \textcolor{gray}{55.6}\,{\scriptsize\textcolor{gray}{$\pm$1.6}} & 65.1\,{\scriptsize\textcolor{gray}{$\pm$1.5}} & \underline{65.3}\,{\scriptsize\textcolor{gray}{$\pm$0.7}} & \underline{65.3}\,{\scriptsize\textcolor{gray}{$\pm$0.8}} & 64.7\,{\scriptsize\textcolor{gray}{$\pm$1.3}} & 64.3\,{\scriptsize\textcolor{gray}{$\pm$0.7}} & \textbf{66.6}\,{\scriptsize\textcolor{gray}{$\pm$0.4}} \\
\cmidrule(lr){1-1}\cmidrule(lr){2-3}\cmidrule(lr){4-9}
\multirow{2}{*}{ALFWorld} \, {\scriptsize Gemma} & \textcolor{gray}{74.5} & \textcolor{gray}{76.6}\,{\scriptsize\textcolor{gray}{$\pm$2.8}} & 74.9\,{\scriptsize\textcolor{gray}{$\pm$1.2}} & \underline{82.0}\,{\scriptsize\textcolor{gray}{$\pm$2.7}} & 81.3\,{\scriptsize\textcolor{gray}{$\pm$1.2}} & 77.4\,{\scriptsize\textcolor{gray}{$\pm$2.5}} & 76.1\,{\scriptsize\textcolor{gray}{$\pm$1.2}} & \textbf{82.6}\,{\scriptsize\textcolor{gray}{$\pm$2.2}} \\
\hphantom{ALFWorld} \, {\scriptsize Qwen} & \textcolor{gray}{52.5} & \textcolor{gray}{69.0}\,{\scriptsize\textcolor{gray}{$\pm$2.8}} & 68.2\,{\scriptsize\textcolor{gray}{$\pm$2.7}} & 67.7\,{\scriptsize\textcolor{gray}{$\pm$5.3}} & \underline{70.1}\,{\scriptsize\textcolor{gray}{$\pm$1.7}} & 62.3\,{\scriptsize\textcolor{gray}{$\pm$2.1}} & 69.4\,{\scriptsize\textcolor{gray}{$\pm$2.3}} & \textbf{71.1}\,{\scriptsize\textcolor{gray}{$\pm$2.6}} \\
\cmidrule(lr){1-1}\cmidrule(lr){2-3}\cmidrule(lr){4-9}
\multirow{2}{*}{WebShop} \, {\scriptsize Gemma} & \textcolor{gray}{66.7} & \textcolor{gray}{68.7}\,{\scriptsize\textcolor{gray}{$\pm$0.9}} & 67.7\,{\scriptsize\textcolor{gray}{$\pm$0.3}} & 69.0\,{\scriptsize\textcolor{gray}{$\pm$0.4}} & \underline{70.1}\,{\scriptsize\textcolor{gray}{$\pm$1.0}} & 67.5\,{\scriptsize\textcolor{gray}{$\pm$0.2}} & 68.6\,{\scriptsize\textcolor{gray}{$\pm$0.5}} & \textbf{72.6}\,{\scriptsize\textcolor{gray}{$\pm$0.4}} \\
\hphantom{WebShop} \, {\scriptsize Qwen} & \textcolor{gray}{62.5} & \textcolor{gray}{63.9}\,{\scriptsize\textcolor{gray}{$\pm$0.7}} & 61.4\,{\scriptsize\textcolor{gray}{$\pm$0.3}} & \textbf{66.0}\,{\scriptsize\textcolor{gray}{$\pm$0.8}} & 64.7\,{\scriptsize\textcolor{gray}{$\pm$1.8}} & 64.0\,{\scriptsize\textcolor{gray}{$\pm$0.6}} & 64.2\,{\scriptsize\textcolor{gray}{$\pm$0.0}} & \underline{65.6}\,{\scriptsize\textcolor{gray}{$\pm$0.8}} \\
\bottomrule\end{tabular}
\end{table}

The most direct way to turn uncertainty into a steering policy is to threshold the signal that best detects failures, as uncertainty-triggered methods commonly do~\citep{jiang2023active,varshney2023stitch,han2024uala,barbi2025preventing,zhang2026agentic,sun2026deep,piatrashyn2026redact}.
Section~\ref{sec:findings} shows why this falls short: such a trigger also fires on successful trajectories, and the best detector is not the best locator.
VoS therefore decides the two questions separately: a \emph{steering value function}, learned from SOT, estimates what steering would achieve at each step, its value of computation in the sense of rational metareasoning~\citep{russell1991principles}, and decides where to steer; a \emph{harm-budgeted trigger} decides whether.
Unlike the controllers of \citet{zhang2026calibration} and DIAL~\citep{li2026same}, which decide at each state in isolation, VoS compares all steps of a trajectory offline and limits how often successful trajectories are disturbed; unlike DIAL, it learns from the size of the effect rather than from a binary label.

\subsection{Steering Value and Harm-Budgeted Triggering}
\label{sec:learning}

The value of steering $\tau$ at step $t$ with mechanism $m$ is how much steering there is expected to change the final task score, given the uncertainty scores that the monitor observes: $V_m(\tau, t) = \mathrm{E}\,[\Delta_m(\tau, t) \mid u_1, \dots, u_T]$, with $T = N$ offline and $T = t$ online.
It is the advantage of steering over not steering~\citep{sutton2018reinforcement}.
The monitor steers at the step of highest estimated value $\hat{V}_m$ if this value reaches a threshold $\theta$:
\begin{equation}
    \pi(\tau) =
    \begin{cases}
        (t^{\star}, m) & \text{if } \hat{V}_m(\tau, t^{\star}) \ge \theta, \\
        \emptyset & \text{otherwise,}
    \end{cases}
    \qquad
    t^{\star} = \arg\max_{t < N} \hat{V}_m(\tau, t).
    \label{eq:monitor}
\end{equation}
Online, where the estimates arrive one step at a time, it steers at the first step with $\hat{V}_m(\tau, t) \ge \theta$.

\begin{figure}[t]
\centering
\includegraphics[width=0.99\linewidth]{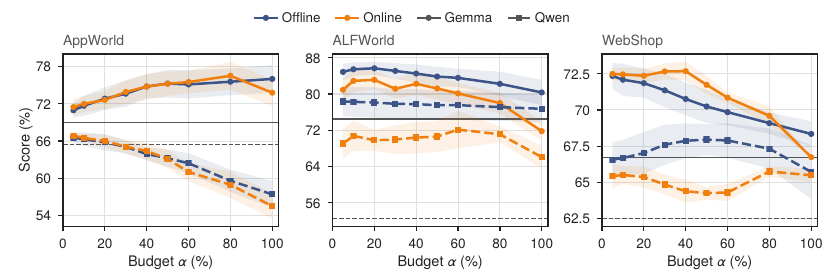}
\caption{Final score of VoS against the budget $\alpha$, on each benchmark's own metric in \%, for Gemma (solid) and Qwen (dashed). Grey lines: unmodified execution; bands: $\pm 1$ standard deviation over the splits.}
\label{fig:results}
\end{figure}

\par\textbf{Steering value function.}\quad
Because $\mathcal{D}$ measures the outcome of every step and mechanism, not only of the steps that some trigger selected, we fit $\hat{V}_m = f_m(x_t)$ by regression on $\mathcal{D}$ without off-policy correction~\citep{dudik2011doubly,bottou2013counterfactual}.
Failure labels could not supervise it: they do not record where, or whether, steering would have changed the result, and Section~\ref{sec:where} shows that the relation of a signal to failure differs from its relation to the value of steering.
The features $x_t$ describe every signal relative to the trajectory offline and to the prefix online, together with the step index, so that the regression can use each signal both to tell trajectories apart and to tell their steps apart; Appendix~\ref{app:monitor} gives the details.
Since the mechanisms disagree on where to steer, as Section~\ref{sec:which} shows, $f_m$ is fitted per mechanism; VoS uses memory-based guidance, the most robust mechanism on both agents in Figures~\ref{fig:s3}a and~\ref{fig:s3_qwen}a, and we test below whether choosing the mechanism as well improves on it.

\par\textbf{Harm-budgeted triggering.}\quad
Section~\ref{sec:whether} shows that uncertainty detects failing trajectories only imperfectly and that every trigger on a successful trajectory can only lower its score, so the monitor limits harm in two ways.
First, $f_m$ is trained on successful as well as failed trajectories, so that their negative outcomes are part of its target.
Second, a budget $\alpha$ on the false-trigger rate sets the threshold, a criterion of the Neyman--Pearson type~\citep{scott2005neyman}: the monitor should recover as much as possible on failed trajectories while steering at most a fraction $\alpha$ of successful ones.
$\theta_{\alpha}$ is the $(1-\alpha)$ quantile of the largest out-of-fold estimate $\max_t \hat{V}_m(\tau, t)$ of the successful training trajectories~\citep{angelopoulos2024conformal}, and $\alpha$ is chosen between $5\%$ and $30\%$ as the budget with the highest task score on the training trajectories, as detailed in Appendix~\ref{app:monitor}.
The budget is calibrated, not guaranteed: the test trajectories are scored by a model refitted on all training trajectories, not by the out-of-fold models that set $\theta_{\alpha}$.

\subsection{Main Results}
\label{sec:results}

\par\textbf{Policies and protocol.}\quad
We compare with five uncertainty-triggered methods: AUQ~\citep{zhang2026agentic} thresholds self-evaluated confidence, DIAL~\citep{li2026same} and \citet{barbi2025preventing} learn from the step index and token entropy, DS-MCM~\citep{sun2026deep} flags deviating reasoning entropy, and ReDAct~\citep{piatrashyn2026redact} thresholds perplexity.
We re-implement only their triggers on SOT, as Appendix~\ref{app:baselines} details, so that all methods decide on the same trajectories.
As a reference, \emph{always steer} steers every trajectory once, at a step chosen on the training half.
All policies are evaluated on the ten splits of Appendix~\ref{app:setup}.

\par\textbf{VoS leads where steering every trajectory is a gamble.}\quad
VoS has the highest score in $11$ of the $12$ settings in Table~\ref{tab:main}, also when every method steers with memory-based guidance in Table~\ref{tab:main_mem}, and remains the best policy in $10$ in Table~\ref{tab:net_resampling}, which nets out the change that re-running from the same step already produces.
Always steering is a gamble: on AppWorld it helps Gemma but costs Qwen $9.9$ points, whereas VoS gains more on Gemma and does not lose in any setting.
Steering almost every trajectory also hurts VoS itself: raising its budget to $100\%$ lowers its gain in nine of the twelve settings in Figure~\ref{fig:results}.
\textbf{Whether to steer therefore matters as much as where}, and the balance between recovery and harm differs between settings, which is why the budget is chosen on training trajectories.

\par\textbf{The gain comes from uncertainty, used through a learned value function.}\quad
On AppWorld, DIAL and \citet{barbi2025preventing}, which see only the step index and token entropy, gain at most $1.0$ point, so the gain of VoS comes from the broader set of uncertainty signals.
Thresholding a single signal does not suffice either: AUQ, DS-MCM, and ReDAct gain at most half as much as VoS on both agents, although AUQ thresholds self-evaluation, the strongest family in Figure~\ref{fig:ablation}.

\par\textbf{Routing does not improve on a fixed mechanism.}\quad
In Table~\ref{tab:mechanisms}, choosing at every step the mechanism of highest estimated value does not improve on memory-based guidance alone, and monitors on any other single mechanism are worse: uncertainty indicates which trajectories to steer, but not which mechanism to use at a step.

\subsection{What the Monitor Relies On}
\label{sec:ablation}

\begin{figure}[t]
\centering
\begin{minipage}[t]{0.49\linewidth}\vspace{0pt}\centering
\includegraphics[height=97.5pt]{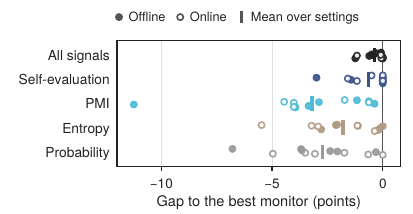}\\[1pt]{\small\phantom{(a)}}
\caption{Gap of each monitor, trained on all signals or on one family, to the best of the five in each of the twelve settings.}
\label{fig:ablation}
\end{minipage}\hfill
\begin{minipage}[t]{0.49\linewidth}\vspace{0pt}\centering
\begin{minipage}[b]{0.53\linewidth}\centering
\includegraphics[height=97.5pt]{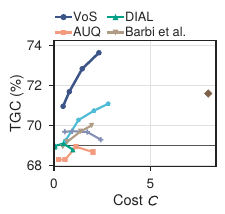}\\[1pt]{\small (a)}
\end{minipage}\hfill
\begin{minipage}[b]{0.46\linewidth}\centering
\includegraphics[height=97.5pt]{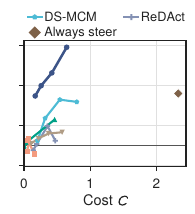}\\[1pt]{\small (b)}
\end{minipage}
\caption{Task goal completion against cost for budgets $\alpha$ from $5\%$ to $30\%$ (AppWorld, Gemma; line: unmodified execution).}
\label{fig:cost}
\end{minipage}
\end{figure}

\par\textbf{No single family is reliable; learning from all of them is.}\quad
Among the monitors trained on one family alone in Figure~\ref{fig:ablation}, self-evaluation is the best single family in nine of the twelve settings but not in the other three, and the token probabilities that uncertainty-triggered methods commonly use fall short of the best monitor by $2.7$ points on average.
The monitor on all fourteen estimators is never more than $1.2$ points below the best one.
A black-box signal is no substitute: the confidence that the agent verbalizes when asked~\citep{tian2023calibration}, the signal behind AUQ, is $1.0$ at $86\%$ of Gemma's steps, so a monitor on it alone gains at most about a third of what VoS gains on AppWorld.
Which uncertainty signal carries the value of steering depends on the benchmark and the agent, so the monitor learns from all of them.

\setlength{\columnsep}{8pt}\begin{wraptable}{r}{0.42\linewidth}
\vspace{-\intextsep}\vspace{-\abovecaptionskip}
\centering
\small
\setlength{\tabcolsep}{5pt}
\caption{VoS trained on the targets of prior work: task goal completion in \% on AppWorld with Gemma.}
\label{tab:supervision}
\vspace{3pt}
\begin{tabular}{@{}lcc@{}}
\toprule
Target & Offline & Online \\
\midrule
Failure label & $69.3$ & $72.5$ \\
DIAL label & $70.8$ & $72.1$ \\
PRM, labelled & $71.6$ & $72.6$ \\
PRM, MC rollout & $71.4$ & $72.1$ \\
Ours & $\mathbf{73.4}$ & $\mathbf{73.6}$ \\
\bottomrule
\end{tabular}

\vspace{-0.5\intextsep}
\end{wraptable}
\par\textbf{Supervision from prior work.}\quad
Table~\ref{tab:supervision} exchanges only the training target of the monitor for the supervision of prior work: trajectory failure~\citep{barbi2025preventing,zhang2026agentic}, DIAL's binary label of whether steering helps~\citep{li2026same}, step correctness judged in hindsight~\citep{lightman2024verify,zhang2025which}, and the Monte-Carlo value of continuing without intervention~\citep{wang2024mathshepherd}.
None matches the measured advantage offline: failure labels recover least, the binary label misses the size of the effect, and step correctness selects the wrong step, since the judged first error lies in the last third of a recoverable trajectory and the most helpful steps in the middle, and steering at a step judged incorrect even gains less than at one judged correct.
Locating the error is not the same problem as locating the step at which steering helps, as Section~\ref{sec:where} anticipated.

\subsection{What Steering Costs}
\label{sec:cost}

\begin{figure}[t]
\centering
\begin{minipage}[c]{0.5\linewidth}
\includegraphics[width=\linewidth]{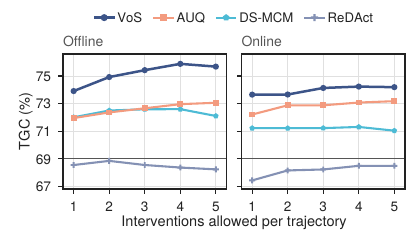}
\end{minipage}\hfill
\begin{minipage}[c]{0.46\linewidth}
\caption{Task goal completion against the number of steering events allowed, in real runs (AppWorld, Gemma; line: unmodified execution).}
\label{fig:multi}
\end{minipage}
\end{figure}
\par\textbf{VoS buys more gain for less.}\quad
Over the range of cost that its curve spans in Figure~\ref{fig:cost}, VoS lies above every method from prior work, which gain almost nothing at lower cost, and at less than a third of the cost it exceeds the best score that any of them attains at \emph{any} budget.
Counting tokens gives the same picture in Figure~\ref{fig:tokens}, and the mechanism's own call, which $C$ includes, is at most $11\%$ of a steering event on Gemma and changes no ordering among the methods in Table~\ref{tab:mech_cost}.

\par\textbf{Steering more than once adds modestly, and VoS stays ahead.}\quad
In real runs with up to five steering events per trajectory, shown in Figure~\ref{fig:multi}, a single event already yields at least $70\%$ of the best gain of every method that gains.
VoS is the best at every number of steering events, so restricting the evaluation to one event per trajectory does not favor it.

\section{Related Work}
\label{sec:related}

\par\textbf{Uncertainty quantification and uncertainty as a trigger.}\quad
Uncertainty estimators for LLMs score a generation by its token probabilities~\citep{fomicheva2020unsupervised,malinin2021uncertainty}, by the consistency of sampled generations~\citep{kuhn2023semantic,farquhar2024semantic,manakul2023selfcheckgpt,lin2024generating}, by its internal representations~\citep{ren2023ood}, or by the model's own confidence, verbalized or scored~\citep{lin2022teaching,tian2023calibration,xiong2024can,kadavath2022language}, and benchmarks such as LM-Polygraph evaluate them by how well they detect errors~\citep{fadeeva2023lmpolygraph,vashurin2025benchmarking}.
Uncertainty also triggers retrieval or verification during generation~\citep{jiang2023active,varshney2023stitch,su2024dragin}, tool calls~\citep{han2024uala}, and requests for help~\citep{ren2023robots} or clarification~\citep{hu2024uncertainty,suri2026structured}.
In agents, where it propagates across steps~\citep{zhao2025saup,duan2025uprop}, it triggers communication resets~\citep{barbi2025preventing}, reflection~\citep{zhang2026agentic,sun2026deep}, and deferral to a stronger model~\citep{piatrashyn2026redact}, and these methods are evaluated by the gain of the whole pipeline.
We evaluate $14$ such estimators, from four families, by what each contributes to the decisions of whether, where, and with which mechanism to steer, measured at every step.

\par\textbf{Steering mechanisms and deferral.}\quad
An agent can be corrected by self-reflection~\citep{shinn2023reflexion,madaan2023selfrefine,kim2025reflact}, tool-based critique~\citep{gou2024critic}, or past experience~\citep{zhao2024expel,wang2024awm}, although self-correction without reliable feedback can degrade results~\citep{huang2024cannot,kamoi2024when}; a query can instead be deferred to a stronger model~\citep{chen2023frugalgpt,yue2024cascades,aggarwal2024automix,ong2025routellm,mozannar2020consistent} or withheld, as in selective prediction~\citep{elyaniv2010foundations,geifman2017selective}.
The four mechanisms of Section~\ref{sec:mechanisms} follow these works.

\par\textbf{Deciding when to intervene.}\quad
Accurate failure prediction does not by itself make intervention effective~\citep{vasudev2026accurate,zhang2026calibration}, and Section~\ref{sec:findings} shows why for uncertainty.
DIAL~\citep{li2026same} and \citet{zhang2026calibration} judge interventions by their counterfactual outcomes, but only at the states that a trigger selects, whereas SOT measures them at every step, so that a monitor can compare the steps of a trajectory.
Process reward models~\citep{lightman2024verify,wang2024mathshepherd} and failure attribution~\citep{zhang2025which} locate the erroneous step instead, which, as Section~\ref{sec:ablation} shows, need not be the step at which steering helps most.
Off-policy evaluation estimates counterfactual outcomes from logged decisions~\citep{dudik2011doubly,bottou2013counterfactual}; SOT observes them by re-executing the agent.

\section{Conclusion}
\label{sec:conclusion}

We asked whether uncertainty can guide effective steering of LLM agents, and measured in SOT what steering achieves at every step of every trajectory with four mechanisms.
Uncertainty identifies failing trajectories, but used directly, today's estimators are an inefficient guide to multi-step steering: they were designed to detect errors in an output, and no single one of them reliably locates the step at which steering helps or tells which mechanism to use.
In our experiments, their value for steering is realized by learning from measured outcomes rather than by thresholding: VoS learns a value function over all signals from SOT, pairs it with a harm-budgeted trigger, and improves on unmodified execution in all $12$ settings.
This learning, however, needs counterfactual outcomes for each benchmark, agent, and mechanism.
A natural next step is uncertainty quantification designed for steering multi-step agents, with estimators whose target is the value of steering at a step rather than the correctness of an output, so that a monitor needs less counterfactual supervision.

\par\textbf{Limitations.}\quad
SOT is expensive to build, since it re-runs the agent from every step with every mechanism.
Our estimators need the agent's token distributions, so we study only open-weight agents of moderate size; black-box signals such as verbalized confidence are much less effective.

\raggedbottom

\bibliography{references}
\bibliographystyle{preprint}

\clearpage
\appendix
\makeatletter\setlength{\@fptop}{0pt}\setlength{\@fpsep}{\floatsep}\setlength{\@fpbot}{0pt plus 1fil}\makeatother
\section{Experimental Setup}
\label{app:setup}

Table~\ref{tab:setup} lists the benchmarks and the settings of the agents.

\begin{table}[!htbp]
\caption{Benchmarks and agent settings.}
\label{tab:setup}
\centering
\small
\begin{tabular}{@{}l >{\raggedright\arraybackslash}p{0.25\linewidth} >{\raggedright\arraybackslash}p{0.2\linewidth} >{\raggedright\arraybackslash}p{0.25\linewidth}@{}}\toprule
& AppWorld & ALFWorld & WebShop\\\midrule
Tasks & 258 & 274 & 400\\\addlinespace
Note built from & failed train tasks & 150 train games & failed train goals\\\addlinespace
Action & one Python code block calling app APIs & one text command & \texttt{search[\dots]} or \texttt{click[\dots]}\\\midrule
Decoding & \multicolumn{3}{l}{temperature 0, seed 100}\\
\bottomrule\end{tabular}
\end{table}

\par\textbf{Splits and training data.}\quad
Steering policies are trained and evaluated on SOT, and everything they need---models, thresholds, and the budget $\alpha$---is chosen on training data only.
On AppWorld and ALFWorld, the table covers the evaluated tasks only, and every policy is trained and evaluated on ten random splits of these tasks into halves, stratified by failure.
On WebShop, it also covers the 300 training goals, so a policy is trained on them and evaluated on the 100 held-out goals, with deviations over ten draws of 250 training goals.
The memory note is synthesized from failed reference trajectories of each benchmark's training split, never from an evaluated task; on AppWorld and ALFWorld, these tasks carry no outcomes and are disjoint from both halves.

\section{Steering Mechanisms}
\label{app:mechanisms}

The boxes give the prompts of the four mechanisms on AppWorld, abridged ([\dots]): the prompt to the model that writes the message, and the message inserted into the agent's context.

\begin{promptbox}{Reflection}
\promptlabel{Prompt to the agent's model}
Your task: \promptvar{instruction}\\
What you have done so far, most recent last: \promptvar{last 40 steps}\\
Reflect, in the first person and in at most four sentences, on your current state in relation to the task goal. (1) State what the most recent output showed [\dots]. (2) List what the task goal requires, item by item, and for each item say whether it is confirmed by an output above, held only as an assumption, or not yet obtained. [\dots] (4) End by naming the single first gap to close next. [\dots] Do not write code. Begin your reply with "Reflection:" and write nothing else.
\tcblower
\promptlabel{Message to the agent}
Pause. Below is your reflection on your current state in relation to the task goal, followed by how to reason for the rest of this task.\\
Reflection: \promptvar{reflection}\\
Your next code block must act on the first gap this reflection names [\dots]. From now on, at every turn, before writing any code, first reflect on your current state in relation to the task goal, in at most four sentences, and only then write the code for this turn. [\dots]
\end{promptbox}

\begin{promptbox}{Critic feedback}
\promptlabel{Brief to the critic, an agent in a copy of the environment at step $t$}
STOP pursuing the original task. New standing instruction: verify the work above before it goes any further. [\dots] What's the problem with the trajectory above? Check it against the live app state with the tools before you judge it [\dots]. Take as many steps as you need, but only read; change nothing. [\dots] Then write a plain-prose critique [\dots]: 1. Requirements [\dots] 2. Truthfulness [\dots] 3. Blocker [\dots] 4. Problem [\dots] Finish with exactly one line: VERDICT: correct [\dots], or VERDICT: incorrect otherwise. If the verdict is incorrect, then add the line "Here's a better next step:" and exactly one code block [\dots].
\tcblower
\promptlabel{Message to the agent}
Your work so far has been checked against the live app state. The critique and a corrected next step are below. Read the critique, run the corrected step as your next action, then continue working on the original task.\\
{}[Critique]: \promptvar{critique}\\
{}[Corrected next step]: \promptvar{code}
\end{promptbox}

\begin{promptbox}{Stronger-model assistance}
\promptlabel{Prompt to DeepSeek-V4-Flash (a second call turns the reasoning into one code block)}
TASK DESCRIPTION: \promptvar{instruction}\\
ENVIRONMENT HISTORY: \promptvar{last 40 steps}\\
AVAILABLE ACTIONS: \promptvar{one Python code block that calls app APIs [\dots]}\\
Think step by step, but keep your reasoning short, precise, and grounded only in the history above. [\dots] 1. Summarize only the important facts the outputs establish [\dots] 2. Identify the current subgoal [\dots] 3. Check the last few actions for useless repetition [\dots] 4. Choose the one step that makes the most direct progress toward the current subgoal. [\dots] After forming your initial reasoning, perform a brief self-critique [\dots]
\tcblower
\promptlabel{Message to the agent}
Your next action has been decided for you. Execute exactly the code below as your next step, without changes, then continue working on the original task from its output.\\
{}[Supervisor reasoning]: \promptvar{reasoning}\\
{}[Deferred action]: \promptvar{code}
\end{promptbox}

\begin{promptbox}{Memory-based guidance}
\promptlabel{Prompt to the agent's model as advisor}
An agent is working on this task: \promptvar{instruction}\\
Here is what it has done so far, most recent last: \promptvar{last 40 steps}\\
Here are failure patterns this agent has shown on other tasks [\dots]: \promptvar{note}\\
{}[\dots] Decide whether any pattern applies to what this agent has just done. Then write the guidance it should receive, in this shape and nothing else:\\
DIAGNOSIS: one sentence on what is wrong or missing right now. [\dots]\\
NEXT: the single concrete next action to take, naming the exact API call or computation. [\dots]\\
THEN: at most two further actions after that.
\tcblower
\promptlabel{Message to the agent}
Guidance on your progress so far is below. Follow it as your next action, then continue working on the original task.\\
DIAGNOSIS: \promptvar{\dots}\quad NEXT: \promptvar{\dots}\quad THEN: \promptvar{\dots}\\
Keep these in mind for the rest of the task: \promptvar{note}
\end{promptbox}

\section{Details of the Monitor}
\label{app:monitor}

\par\textbf{Uncertainty estimators.}\quad
The monitor uses the 14 estimators of Section~\ref{sec:data}, computed as defined in LM-Polygraph~\citep{fadeeva2023lmpolygraph}, in four families: probability (maximum sequence probability, maximum token probability, and perplexity); entropy (mean and maximum token entropy, self-certainty, Fisher--Rao distance, and R\'enyi negentropy); pointwise mutual information (PMI and conditional PMI, each as the mean and the maximum over the tokens of the step); and self-evaluation ($P(\mathrm{True})$ and its normalized form $p(\mathrm{True}) / (p(\mathrm{True}) + p(\mathrm{False}))$).
Table~\ref{tab:uq_cost} lists the additional computation that each family requires.
A value that is missing at a step is replaced by the mean of the estimator over the same trajectory.

\par\textbf{Features.}\quad
The features $x_t$ combine all signals with the step index.
In the offline setting they describe each signal at step $t$ relative to the complete trajectory, through its value, its rank and standardized value, its change at the next step, and summary statistics over all steps; in the online setting they describe it relative to the prefix, through its current value, running statistics, percentile, and change from the previous step.
Offline, two of these forms correspond to the two abilities of Figure~\ref{fig:s3}b: the summary statistics, which are the same at every step and therefore only tell trajectories apart (detection), and the rank and standardized value within the trajectory, which tell its steps apart (localization); the regression learns how much each form of each signal counts.

\par\textbf{Learners and hyperparameters.}\quad
The value function is an ensemble of gradient-boosted regression trees~\citep{friedman2001greedy,ke2017lightgbm} in the offline setting and a ridge regression~\citep{hoerl1970ridge} in the online setting, both as implemented in scikit-learn~\citep{pedregosa2011scikit}, with the hyperparameters in Table~\ref{tab:monitor_hparams}.
The hyperparameters were chosen once, on AppWorld: every configuration of the grids was scored by the out-of-fold task score on the training halves of the ten splits at a fixed budget ($\alpha = 30\%$ for Gemma and $10\%$ for Qwen), and the configuration that scored best on both agents together was kept fixed for all splits, both agents, and all three benchmarks; ALFWorld and WebShop were not tuned.
The random seed of the learner is the index of the split.

\begin{table}[!htbp]
\caption{Learners and hyperparameters of the monitor, and the grids from which the hyperparameters were chosen.}
\label{tab:monitor_hparams}
\centering
\small
\begin{tabular}{@{}l >{\raggedright\arraybackslash}p{0.47\linewidth} >{\raggedright\arraybackslash}p{0.28\linewidth}@{}}\toprule
& Offline & Online\\\midrule
Model & Gradient-boosted regression trees (\texttt{HistGradientBoostingRegressor}) & Ridge regression after mean imputation and standardization\\\addlinespace
Hyperparameters & At most 3 leaves per tree, learning rate 0.05, 200 iterations, at least 20 samples per leaf, $\ell_2$ regularization 1.0, no early stopping & $\lambda = 300$\\\addlinespace
Grid & 18 configurations: 3 to 16 leaves, learning rate 0.03 to 0.1, 100 to 400 iterations, 20 or 40 samples per leaf, $\ell_2$ regularization 0 to 5 & 7 values, $\lambda \in \{10, 30, 100, \dots, 10^4\}$\\
\bottomrule\end{tabular}
\end{table}

\par\textbf{Threshold calibration.}\quad
Within the training half, a four-fold cross-validation grouped by task gives an out-of-fold estimate $\hat{V}_m(\tau, t)$ at every step of every training trajectory.
For each successful training trajectory we take the largest out-of-fold estimate over its candidate steps, and $\theta_\alpha$ is the $(1-\alpha)$ quantile of these maxima.
The budget is chosen from $\alpha \in \{5, 10, 20, 30\}\%$ by running the policy with the out-of-fold estimates and $\theta_\alpha$ on the training half and keeping the $\alpha$ of highest task score; the model that scores the test half is then refitted on the whole training half.
The candidate steps are the non-terminal steps $t < N$.
Offline, the monitor steers at the step of largest estimate if this estimate reaches $\theta_\alpha$; online, it steers at the first step $t \ge t_{\min} = 3$ whose estimate reaches $\theta_\alpha$, and the calibration uses only these steps.
The budgets from $40\%$ to $100\%$ in Figure~\ref{fig:results} are evaluated only to draw the curves and are never selected.

\par\textbf{Training target and weights.}\quad
A separate value function $f_m$ is fitted for each mechanism $m$ on all candidate steps of the failed and the successful training trajectories, with the outcome $\Delta_m(\tau, t)$ of Eq.~\ref{eq:outcome} as the target; on AppWorld, the outcome is computed from the fraction of tests passed.
Each step is weighted by $1/N_\tau$, so that long and short trajectories contribute the same total weight to the loss.

\section{Methods from Prior Work}
\label{app:baselines}

We re-implement the triggering rule of each method on SOT; like the threshold of VoS, its threshold is set by the budget calibration of Section~\ref{sec:learning} on the training half.

\par\textbf{AUQ.}\quad
AUQ~\citep{zhang2026agentic} triggers reflection when the agent's confidence is low, and we keep this trigger.
Its uncertainty-aware memory, which has the agent write down at every step its confidence and what it is uncertain about, is not reproduced: in our runs it doubles the agent's output tokens and no task gets stuck in a loop, so it acts as a prompt that makes the agent think one step further rather than as uncertainty that decides when to steer.
Its consistency-weighted best-of-$N$ reflection is a steering mechanism rather than a trigger and is not reproduced either.

\par\textbf{DIAL.}\quad
DIAL~\citep{li2026same} decides when to steer with an $\ell_1$-regularized logistic gate trained on a binary label of whether steering helps, and we keep this gate.
The optimizer that the gate invokes is not reproduced: it ranks candidate actions by the environment's reward, with access to the environment's scoring function (Appendix D.1 of the original), which on AppWorld would mean scoring candidate actions with the hidden unit tests.
No other method has this information, and an optimizer without it would no longer be the original, so DIAL steers with memory-based guidance.

\par\textbf{Barbi et al.}\quad
\citet{barbi2025preventing} predict task success with a polynomial ridge classifier, and we keep this classifier.
Their intervention resets the communication between the agents of a multi-agent system; our agents act alone, so there is no communication to reset, and we steer with memory-based guidance instead.

\par\textbf{DS-MCM.}\quad
DS-MCM~\citep{sun2026deep} triggers when the reasoning entropy of a step deviates from its level on successful steps; we keep this trigger and steer with memory-based guidance.

\par\textbf{ReDAct.}\quad
ReDAct~\citep{piatrashyn2026redact} defers a step to a stronger model when its perplexity is high, and we keep this deferral as it is.

\section{Additional Results}
\label{app:results}

Table~\ref{tab:main_mem} repeats Table~\ref{tab:main} with memory-based guidance for every policy; Table~\ref{tab:mechanisms} compares the monitors built on each mechanism with the routed monitor of Section~\ref{sec:results}, and Table~\ref{tab:mech_cost} gives the cost of the mechanisms' own calls used in Section~\ref{sec:cost}.

\begin{table}[!htbp]
\caption{Table~\ref{tab:main} with memory-based guidance for every policy: AUQ and ReDAct steer with it instead of reflection and the stronger model, so that all methods differ only in when and where they steer. Gray: references, not ranked; bold and underline: best and second-best policy in each row.}
\label{tab:main_mem}
\centering
\footnotesize
\setlength{\tabcolsep}{2.6pt}\renewcommand{\arraystretch}{1.08}
\begin{tabular}{@{}l cc ccccc >{\columncolor{ourscol}}c@{}}\toprule
& \multicolumn{2}{c}{\textcolor{gray}{References}} & \multicolumn{6}{c}{Steering policies (all with memory-based guidance)}\\
\cmidrule(lr){2-3}\cmidrule(lr){4-9}
& \textcolor{gray}{Unmod.} & \textcolor{gray}{Always} & AUQ & DIAL & Barbi et al. & DS-MCM & ReDAct & VoS (ours)\\\midrule
\multicolumn{9}{@{}l}{\emph{Offline}}\\[1pt]
\multirow{2}{*}{AppWorld} \, {\scriptsize Gemma} & \textcolor{gray}{69.0} & \textcolor{gray}{71.6}\,{\scriptsize\textcolor{gray}{$\pm$1.5}} & \underline{72.2}\,{\scriptsize\textcolor{gray}{$\pm$1.6}} & 68.9\,{\scriptsize\textcolor{gray}{$\pm$0.3}} & 69.7\,{\scriptsize\textcolor{gray}{$\pm$1.2}} & 70.4\,{\scriptsize\textcolor{gray}{$\pm$0.9}} & 69.5\,{\scriptsize\textcolor{gray}{$\pm$0.6}} & \textbf{73.4}\,{\scriptsize\textcolor{gray}{$\pm$1.9}} \\
\hphantom{AppWorld} \, {\scriptsize Qwen} & \textcolor{gray}{65.5} & \textcolor{gray}{55.6}\,{\scriptsize\textcolor{gray}{$\pm$1.6}} & 64.3\,{\scriptsize\textcolor{gray}{$\pm$0.7}} & \underline{65.4}\,{\scriptsize\textcolor{gray}{$\pm$0.4}} & 65.3\,{\scriptsize\textcolor{gray}{$\pm$0.7}} & 64.8\,{\scriptsize\textcolor{gray}{$\pm$0.6}} & 65.3\,{\scriptsize\textcolor{gray}{$\pm$0.5}} & \textbf{66.5}\,{\scriptsize\textcolor{gray}{$\pm$0.8}} \\
\cmidrule(lr){1-1}\cmidrule(lr){2-3}\cmidrule(lr){4-9}
\multirow{2}{*}{ALFWorld} \, {\scriptsize Gemma} & \textcolor{gray}{74.5} & \textcolor{gray}{76.6}\,{\scriptsize\textcolor{gray}{$\pm$2.8}} & 73.6\,{\scriptsize\textcolor{gray}{$\pm$0.8}} & \underline{80.1}\,{\scriptsize\textcolor{gray}{$\pm$1.8}} & 80.0\,{\scriptsize\textcolor{gray}{$\pm$1.8}} & 78.4\,{\scriptsize\textcolor{gray}{$\pm$1.7}} & 77.4\,{\scriptsize\textcolor{gray}{$\pm$1.1}} & \textbf{85.2}\,{\scriptsize\textcolor{gray}{$\pm$1.6}} \\
\hphantom{ALFWorld} \, {\scriptsize Qwen} & \textcolor{gray}{52.5} & \textcolor{gray}{69.0}\,{\scriptsize\textcolor{gray}{$\pm$2.8}} & 55.3\,{\scriptsize\textcolor{gray}{$\pm$1.7}} & 62.0\,{\scriptsize\textcolor{gray}{$\pm$3.5}} & 61.0\,{\scriptsize\textcolor{gray}{$\pm$2.4}} & 57.0\,{\scriptsize\textcolor{gray}{$\pm$1.6}} & \underline{63.5}\,{\scriptsize\textcolor{gray}{$\pm$2.6}} & \textbf{78.4}\,{\scriptsize\textcolor{gray}{$\pm$3.2}} \\
\cmidrule(lr){1-1}\cmidrule(lr){2-3}\cmidrule(lr){4-9}
\multirow{2}{*}{WebShop} \, {\scriptsize Gemma} & \textcolor{gray}{66.7} & \textcolor{gray}{68.7}\,{\scriptsize\textcolor{gray}{$\pm$0.9}} & \underline{71.3}\,{\scriptsize\textcolor{gray}{$\pm$0.8}} & 68.3\,{\scriptsize\textcolor{gray}{$\pm$0.5}} & 69.4\,{\scriptsize\textcolor{gray}{$\pm$0.2}} & 67.6\,{\scriptsize\textcolor{gray}{$\pm$0.3}} & 67.7\,{\scriptsize\textcolor{gray}{$\pm$0.4}} & \textbf{72.3}\,{\scriptsize\textcolor{gray}{$\pm$1.1}} \\
\hphantom{WebShop} \, {\scriptsize Qwen} & \textcolor{gray}{62.5} & \textcolor{gray}{63.9}\,{\scriptsize\textcolor{gray}{$\pm$0.7}} & 63.5\,{\scriptsize\textcolor{gray}{$\pm$0.5}} & 63.3\,{\scriptsize\textcolor{gray}{$\pm$0.6}} & 63.6\,{\scriptsize\textcolor{gray}{$\pm$0.4}} & \underline{65.0}\,{\scriptsize\textcolor{gray}{$\pm$0.0}} & 64.3\,{\scriptsize\textcolor{gray}{$\pm$1.0}} & \textbf{67.1}\,{\scriptsize\textcolor{gray}{$\pm$0.9}} \\
\midrule
\multicolumn{9}{@{}l}{\emph{Online}}\\[1pt]
\multirow{2}{*}{AppWorld} \, {\scriptsize Gemma} & \textcolor{gray}{69.0} & \textcolor{gray}{71.6}\,{\scriptsize\textcolor{gray}{$\pm$1.5}} & \underline{71.9}\,{\scriptsize\textcolor{gray}{$\pm$1.6}} & 70.0\,{\scriptsize\textcolor{gray}{$\pm$1.7}} & 69.7\,{\scriptsize\textcolor{gray}{$\pm$0.8}} & 71.2\,{\scriptsize\textcolor{gray}{$\pm$0.9}} & 70.9\,{\scriptsize\textcolor{gray}{$\pm$1.2}} & \textbf{73.6}\,{\scriptsize\textcolor{gray}{$\pm$1.8}} \\
\hphantom{AppWorld} \, {\scriptsize Qwen} & \textcolor{gray}{65.5} & \textcolor{gray}{55.6}\,{\scriptsize\textcolor{gray}{$\pm$1.6}} & 62.4\,{\scriptsize\textcolor{gray}{$\pm$0.7}} & \underline{65.3}\,{\scriptsize\textcolor{gray}{$\pm$0.7}} & \underline{65.3}\,{\scriptsize\textcolor{gray}{$\pm$0.8}} & 64.7\,{\scriptsize\textcolor{gray}{$\pm$1.3}} & 63.8\,{\scriptsize\textcolor{gray}{$\pm$0.6}} & \textbf{66.6}\,{\scriptsize\textcolor{gray}{$\pm$0.4}} \\
\cmidrule(lr){1-1}\cmidrule(lr){2-3}\cmidrule(lr){4-9}
\multirow{2}{*}{ALFWorld} \, {\scriptsize Gemma} & \textcolor{gray}{74.5} & \textcolor{gray}{76.6}\,{\scriptsize\textcolor{gray}{$\pm$2.8}} & 75.3\,{\scriptsize\textcolor{gray}{$\pm$1.2}} & \underline{82.0}\,{\scriptsize\textcolor{gray}{$\pm$2.7}} & 81.3\,{\scriptsize\textcolor{gray}{$\pm$1.2}} & 77.4\,{\scriptsize\textcolor{gray}{$\pm$2.5}} & 77.9\,{\scriptsize\textcolor{gray}{$\pm$1.3}} & \textbf{82.6}\,{\scriptsize\textcolor{gray}{$\pm$2.2}} \\
\hphantom{ALFWorld} \, {\scriptsize Qwen} & \textcolor{gray}{52.5} & \textcolor{gray}{69.0}\,{\scriptsize\textcolor{gray}{$\pm$2.8}} & 63.8\,{\scriptsize\textcolor{gray}{$\pm$2.2}} & 67.7\,{\scriptsize\textcolor{gray}{$\pm$5.3}} & \underline{70.1}\,{\scriptsize\textcolor{gray}{$\pm$1.7}} & 62.3\,{\scriptsize\textcolor{gray}{$\pm$2.1}} & 65.3\,{\scriptsize\textcolor{gray}{$\pm$3.0}} & \textbf{71.1}\,{\scriptsize\textcolor{gray}{$\pm$2.6}} \\
\cmidrule(lr){1-1}\cmidrule(lr){2-3}\cmidrule(lr){4-9}
\multirow{2}{*}{WebShop} \, {\scriptsize Gemma} & \textcolor{gray}{66.7} & \textcolor{gray}{68.7}\,{\scriptsize\textcolor{gray}{$\pm$0.9}} & 69.8\,{\scriptsize\textcolor{gray}{$\pm$0.3}} & 69.0\,{\scriptsize\textcolor{gray}{$\pm$0.4}} & \underline{70.1}\,{\scriptsize\textcolor{gray}{$\pm$1.0}} & 67.5\,{\scriptsize\textcolor{gray}{$\pm$0.2}} & 68.2\,{\scriptsize\textcolor{gray}{$\pm$0.3}} & \textbf{72.6}\,{\scriptsize\textcolor{gray}{$\pm$0.4}} \\
\hphantom{WebShop} \, {\scriptsize Qwen} & \textcolor{gray}{62.5} & \textcolor{gray}{63.9}\,{\scriptsize\textcolor{gray}{$\pm$0.7}} & 62.7\,{\scriptsize\textcolor{gray}{$\pm$1.3}} & \textbf{66.0}\,{\scriptsize\textcolor{gray}{$\pm$0.8}} & 64.7\,{\scriptsize\textcolor{gray}{$\pm$1.8}} & 64.0\,{\scriptsize\textcolor{gray}{$\pm$0.6}} & 63.9\,{\scriptsize\textcolor{gray}{$\pm$0.0}} & \underline{65.6}\,{\scriptsize\textcolor{gray}{$\pm$0.8}} \\
\bottomrule\end{tabular}
\end{table}

\begin{table}[!htbp]
\caption{Improvement over unmodified execution of the monitor built on each mechanism, and of a routed monitor that chooses at every step the mechanism of highest estimated value (Gemma).}
\label{tab:mechanisms}
\centering
\small
\begin{tabular}{@{}lcccccc@{}}\toprule
& \multicolumn{2}{c}{AppWorld} & \multicolumn{2}{c}{ALFWorld} & \multicolumn{2}{c}{WebShop}\\
\cmidrule(lr){2-3}\cmidrule(lr){4-5}\cmidrule(l){6-7}
& Offline & Online & Offline & Online & Offline & Online\\\midrule
Memory-based guidance (VoS) & $+$.045 & $+$.046 & $+$.107 & $+$.081 & $+$.055 & $+$.058 \\
Reflection & $+$.008 & $+$.010 & $+$.080 & $+$.018 & $+$.026 & $+$.002 \\
Critic & $+$.022 & $+$.002 & $+$.056 & $+$.027 & $+$.051 & $+$.042 \\
Stronger model & $+$.019 & $+$.015 & $+$.055 & $+$.031 & $+$.046 & $+$.037 \\
Routed & $+$.035 & $+$.049 & $+$.092 & $+$.074 & $+$.047 & $+$.055 \\
\bottomrule\end{tabular}
\end{table}

\begin{table}[!htbp]
\caption{Cost of the mechanism's own call per steering event on AppWorld: prompt and completion tokens, number of calls, the tokens in agent-step equivalents, and their share of the total cost of a steering event.}
\label{tab:mech_cost}
\centering
\small
\setlength{\tabcolsep}{4pt}
\begin{tabular}{@{}lcccccccc@{}}\toprule
& \multicolumn{4}{c}{Gemma} & \multicolumn{4}{c}{Qwen}\\
\cmidrule(lr){2-5}\cmidrule(l){6-9}
& Tokens & Calls & Steps & Share & Tokens & Calls & Steps & Share\\\midrule
Reflection & 2{,}261 & 1.00 & 0.30 & 4.0\% & 4{,}258 & 1.00 & 0.48 & 4.3\% \\
Critic & 6{,}300 & 3.02 & 0.84 & 10.5\% & 20{,}094 & 1.06 & 2.28 & 18.8\% \\
Stronger model & 6{,}515 & 2.04 & 0.87 & 9.5\% & 8{,}748 & 2.01 & 0.99 & 8.3\% \\
Memory-based guidance & 3{,}589 & 1.00 & 0.48 & 5.1\% & 6{,}579 & 1.00 & 0.75 & 6.5\% \\
\bottomrule\end{tabular}
\end{table}

\begin{table}[!htbp]
\caption{Change of the benchmark metric net of resampling: for every steered cell, the change produced by a bare rewind at the same step, which continues the trajectory from step $t$ with no message, is subtracted, so a policy is credited only with what its message adds. Metrics as in Table~\ref{tab:main}, as mean $\pm$ standard deviation. The gray reference is a monitor trained on the rewind alone, which learns where re-running helps. Bold and underline: best and second-best policy in each row; shading from red (loss) to green (gain).}
\label{tab:net_resampling}
\centering
\small
\setlength{\tabcolsep}{2.2pt}\renewcommand{\arraystretch}{1.08}
\resizebox{\ifdim\width>\linewidth\linewidth\else\width\fi}{!}{\begin{tabular}{@{}l c ccccc >{\columncolor{ourscol}}c@{}}\toprule
& \textcolor{gray}{Reference} & \multicolumn{6}{c}{Steering policies, net of the rewind}\\
\cmidrule(lr){2-2}\cmidrule(lr){3-8}
& \textcolor{gray}{\begin{tabular}[b]{@{}c@{}}Rewind\\monitor\end{tabular}} & AUQ & DIAL & Barbi et al. & DS-MCM & ReDAct & VoS (ours)\\\midrule
\multicolumn{8}{@{}l}{\emph{Offline}}\\[1pt]
\multirow{2}{*}{AppWorld} \, {\scriptsize Gemma} & \cellcolor[HTML]{DEF0E6}\textcolor{gray}{$+$0.7}{\scriptsize\textcolor{gray}{$\pm$0.7}} & \cellcolor[HTML]{DBEFE4}$+$0.8{\scriptsize\textcolor{gray}{$\pm$1.2}} & \cellcolor[HTML]{D3EBDE}$+$1.2{\scriptsize\textcolor{gray}{$\pm$1.7}} & \cellcolor[HTML]{C5E5D3}$+$2.1{\scriptsize\textcolor{gray}{$\pm$1.3}} & \cellcolor[HTML]{B6DEC7}\underline{$+$3.4}{\scriptsize\textcolor{gray}{$\pm$0.8}} & \cellcolor[HTML]{CBE8D7}$+$1.7{\scriptsize\textcolor{gray}{$\pm$0.8}} & \cellcolor[HTML]{A4D6BA}\textbf{$+$5.2}{\scriptsize\textcolor{gray}{$\pm$1.7}} \\
\hphantom{AppWorld} \, {\scriptsize Qwen} & \cellcolor[HTML]{F2F9F5}\textcolor{gray}{$+$0.1}{\scriptsize\textcolor{gray}{$\pm$1.1}} & \cellcolor[HTML]{FAE3E2}$-$0.6{\scriptsize\textcolor{gray}{$\pm$0.5}} & \cellcolor[HTML]{FBE8E7}$-$0.4{\scriptsize\textcolor{gray}{$\pm$0.5}} & \cellcolor[HTML]{FAE3E2}$-$0.6{\scriptsize\textcolor{gray}{$\pm$0.8}} & \cellcolor[HTML]{F7D4D3}$-$1.4{\scriptsize\textcolor{gray}{$\pm$0.9}} & \cellcolor[HTML]{FDF4F3}\underline{$-$0.1}{\scriptsize\textcolor{gray}{$\pm$0.3}} & \cellcolor[HTML]{D0EADB}\textbf{$+$1.4}{\scriptsize\textcolor{gray}{$\pm$0.7}} \\
\cmidrule(lr){1-1}\cmidrule(lr){2-2}\cmidrule(lr){3-8}
\multirow{2}{*}{ALFWorld} \, {\scriptsize Gemma} & \cellcolor[HTML]{AAD9BE}\textcolor{gray}{$+$4.6}{\scriptsize\textcolor{gray}{$\pm$2.1}} & \cellcolor[HTML]{FDF4F3}$-$0.1{\scriptsize\textcolor{gray}{$\pm$0.9}} & \cellcolor[HTML]{B3DDC5}\underline{$+$3.6}{\scriptsize\textcolor{gray}{$\pm$1.4}} & \cellcolor[HTML]{B7DEC8}$+$3.3{\scriptsize\textcolor{gray}{$\pm$2.0}} & \cellcolor[HTML]{BBE0CB}$+$2.9{\scriptsize\textcolor{gray}{$\pm$1.7}} & \cellcolor[HTML]{F5CDCB}$-$1.9{\scriptsize\textcolor{gray}{$\pm$1.3}} & \cellcolor[HTML]{A6D7BB}\textbf{$+$5.0}{\scriptsize\textcolor{gray}{$\pm$1.7}} \\
\hphantom{ALFWorld} \, {\scriptsize Qwen} & \cellcolor[HTML]{85C8A2}\textcolor{gray}{$+$9.4}{\scriptsize\textcolor{gray}{$\pm$1.9}} & \cellcolor[HTML]{E9F5EE}$+$0.3{\scriptsize\textcolor{gray}{$\pm$0.5}} & \cellcolor[HTML]{93CEAC}\underline{$+$7.4}{\scriptsize\textcolor{gray}{$\pm$5.0}} & \cellcolor[HTML]{A8D8BC}$+$4.8{\scriptsize\textcolor{gray}{$\pm$2.7}} & \cellcolor[HTML]{ABD9BE}$+$4.5{\scriptsize\textcolor{gray}{$\pm$2.2}} & \cellcolor[HTML]{F7D6D4}$-$1.3{\scriptsize\textcolor{gray}{$\pm$1.1}} & \cellcolor[HTML]{81C69F}\textbf{$+$17.3}{\scriptsize\textcolor{gray}{$\pm$4.1}} \\
\cmidrule(lr){1-1}\cmidrule(lr){2-2}\cmidrule(lr){3-8}
\multirow{2}{*}{WebShop} \, {\scriptsize Gemma} & \cellcolor[HTML]{C1E3D0}\textcolor{gray}{$+$2.4}{\scriptsize\textcolor{gray}{$\pm$0.6}} & \cellcolor[HTML]{E9F5EE}$+$0.3{\scriptsize\textcolor{gray}{$\pm$0.1}} & \cellcolor[HTML]{E3F2E9}\underline{$+$0.5}{\scriptsize\textcolor{gray}{$\pm$0.8}} & $+$0.0{\scriptsize\textcolor{gray}{$\pm$0.2}} & \cellcolor[HTML]{E9F5EE}$+$0.3{\scriptsize\textcolor{gray}{$\pm$0.3}} & \cellcolor[HTML]{FBEBEA}$-$0.3{\scriptsize\textcolor{gray}{$\pm$1.2}} & \cellcolor[HTML]{A9D8BD}\textbf{$+$4.7}{\scriptsize\textcolor{gray}{$\pm$1.5}} \\
\hphantom{WebShop} \, {\scriptsize Qwen} & \cellcolor[HTML]{D2EBDC}\textcolor{gray}{$+$1.3}{\scriptsize\textcolor{gray}{$\pm$0.3}} & \cellcolor[HTML]{F6D0CE}$-$1.7{\scriptsize\textcolor{gray}{$\pm$0.4}} & \cellcolor[HTML]{E9F5EE}$+$0.3{\scriptsize\textcolor{gray}{$\pm$1.0}} & \cellcolor[HTML]{D9EEE2}$+$0.9{\scriptsize\textcolor{gray}{$\pm$0.7}} & \cellcolor[HTML]{D0EADB}$+$1.4{\scriptsize\textcolor{gray}{$\pm$0.8}} & \cellcolor[HTML]{C8E6D5}\underline{$+$1.9}{\scriptsize\textcolor{gray}{$\pm$0.7}} & \cellcolor[HTML]{BCE1CC}\textbf{$+$2.8}{\scriptsize\textcolor{gray}{$\pm$1.0}} \\
\midrule
\multicolumn{8}{@{}l}{\emph{Online}}\\[1pt]
\multirow{2}{*}{AppWorld} \, {\scriptsize Gemma} & \textcolor{gray}{$+$0.0}{\scriptsize\textcolor{gray}{$\pm$0.6}} & \cellcolor[HTML]{DBEFE4}$+$0.8{\scriptsize\textcolor{gray}{$\pm$1.5}} & \cellcolor[HTML]{A9D8BD}\underline{$+$4.7}{\scriptsize\textcolor{gray}{$\pm$3.6}} & \cellcolor[HTML]{C3E4D1}$+$2.3{\scriptsize\textcolor{gray}{$\pm$1.6}} & \cellcolor[HTML]{B0DCC3}$+$3.9{\scriptsize\textcolor{gray}{$\pm$0.9}} & \cellcolor[HTML]{C5E5D3}$+$2.1{\scriptsize\textcolor{gray}{$\pm$1.1}} & \cellcolor[HTML]{A5D7BA}\textbf{$+$5.1}{\scriptsize\textcolor{gray}{$\pm$1.3}} \\
\hphantom{AppWorld} \, {\scriptsize Qwen} & \cellcolor[HTML]{FCEFEE}\textcolor{gray}{$-$0.2}{\scriptsize\textcolor{gray}{$\pm$1.0}} & \cellcolor[HTML]{FBEBEA}$-$0.3{\scriptsize\textcolor{gray}{$\pm$0.8}} & \cellcolor[HTML]{FAE6E4}$-$0.5{\scriptsize\textcolor{gray}{$\pm$0.6}} & \cellcolor[HTML]{F9E1E0}$-$0.7{\scriptsize\textcolor{gray}{$\pm$0.6}} & \cellcolor[HTML]{F5CCCA}$-$2.0{\scriptsize\textcolor{gray}{$\pm$1.2}} & \cellcolor[HTML]{F2F9F5}\textbf{$+$0.1}{\scriptsize\textcolor{gray}{$\pm$0.5}} & \cellcolor[HTML]{FDF4F3}\underline{$-$0.1}{\scriptsize\textcolor{gray}{$\pm$0.3}} \\
\cmidrule(lr){1-1}\cmidrule(lr){2-2}\cmidrule(lr){3-8}
\multirow{2}{*}{ALFWorld} \, {\scriptsize Gemma} & \cellcolor[HTML]{CBE8D7}\textcolor{gray}{$+$1.7}{\scriptsize\textcolor{gray}{$\pm$0.9}} & \cellcolor[HTML]{FBEBEA}$-$0.3{\scriptsize\textcolor{gray}{$\pm$0.9}} & \cellcolor[HTML]{BAE0CA}$+$3.0{\scriptsize\textcolor{gray}{$\pm$2.9}} & \cellcolor[HTML]{B0DCC3}\underline{$+$3.9}{\scriptsize\textcolor{gray}{$\pm$1.5}} & \cellcolor[HTML]{C3E4D1}$+$2.3{\scriptsize\textcolor{gray}{$\pm$1.7}} & \cellcolor[HTML]{F8DBD9}$-$1.0{\scriptsize\textcolor{gray}{$\pm$1.2}} & \cellcolor[HTML]{A7D7BC}\textbf{$+$4.9}{\scriptsize\textcolor{gray}{$\pm$2.1}} \\
\hphantom{ALFWorld} \, {\scriptsize Qwen} & \cellcolor[HTML]{CBE8D7}\textcolor{gray}{$+$1.7}{\scriptsize\textcolor{gray}{$\pm$1.8}} & \cellcolor[HTML]{E9F5EE}$+$0.3{\scriptsize\textcolor{gray}{$\pm$0.4}} & \cellcolor[HTML]{81C69F}$+$12.4{\scriptsize\textcolor{gray}{$\pm$3.2}} & \cellcolor[HTML]{81C69F}\underline{$+$13.3}{\scriptsize\textcolor{gray}{$\pm$3.2}} & \cellcolor[HTML]{89CAA5}$+$8.8{\scriptsize\textcolor{gray}{$\pm$1.9}} & \cellcolor[HTML]{C1E3D0}$+$2.4{\scriptsize\textcolor{gray}{$\pm$1.5}} & \cellcolor[HTML]{81C69F}\textbf{$+$15.7}{\scriptsize\textcolor{gray}{$\pm$2.6}} \\
\cmidrule(lr){1-1}\cmidrule(lr){2-2}\cmidrule(lr){3-8}
\multirow{2}{*}{WebShop} \, {\scriptsize Gemma} & \cellcolor[HTML]{BFE2CE}\textcolor{gray}{$+$2.6}{\scriptsize\textcolor{gray}{$\pm$1.2}} & \cellcolor[HTML]{E9F5EE}$+$0.3{\scriptsize\textcolor{gray}{$\pm$0.3}} & \cellcolor[HTML]{FCEFEE}$-$0.2{\scriptsize\textcolor{gray}{$\pm$1.3}} & \cellcolor[HTML]{BCE1CC}\underline{$+$2.8}{\scriptsize\textcolor{gray}{$\pm$1.5}} & \cellcolor[HTML]{F8DBD9}$-$1.0{\scriptsize\textcolor{gray}{$\pm$0.5}} & \cellcolor[HTML]{FCEFEE}$-$0.2{\scriptsize\textcolor{gray}{$\pm$0.9}} & \cellcolor[HTML]{A9D8BD}\textbf{$+$4.7}{\scriptsize\textcolor{gray}{$\pm$0.9}} \\
\hphantom{WebShop} \, {\scriptsize Qwen} & \cellcolor[HTML]{F2F9F5}\textcolor{gray}{$+$0.1}{\scriptsize\textcolor{gray}{$\pm$0.9}} & \cellcolor[HTML]{FBE8E7}$-$0.4{\scriptsize\textcolor{gray}{$\pm$0.6}} & \cellcolor[HTML]{C5E5D3}\textbf{$+$2.1}{\scriptsize\textcolor{gray}{$\pm$0.8}} & \cellcolor[HTML]{CDE8D8}\underline{$+$1.6}{\scriptsize\textcolor{gray}{$\pm$2.3}} & \cellcolor[HTML]{D3EBDE}$+$1.2{\scriptsize\textcolor{gray}{$\pm$1.0}} & \cellcolor[HTML]{DEF0E6}$+$0.7{\scriptsize\textcolor{gray}{$\pm$0.2}} & \cellcolor[HTML]{CDE8D8}\underline{$+$1.6}{\scriptsize\textcolor{gray}{$\pm$1.4}} \\
\bottomrule\end{tabular}}
\end{table}

\section{Additional Figure Details}

\label{app:figure_details}

These figures retain the detailed comparisons behind the main-text summaries, using the same outcomes and task splits.

\begin{figure}[!htbp]
\centering
\begin{minipage}[b]{0.5\linewidth}\centering
\includegraphics[width=\linewidth]{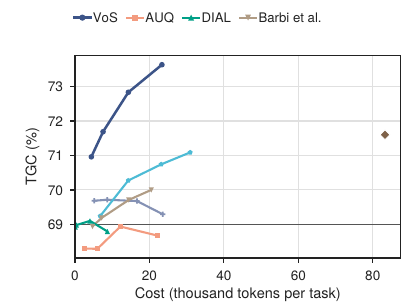}\\[1pt]{\small (a) Offline}
\end{minipage}\hfill
\begin{minipage}[b]{0.5\linewidth}\centering
\includegraphics[width=\linewidth]{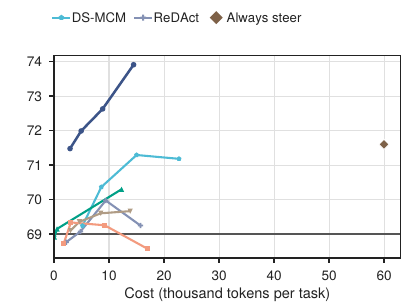}\\[1pt]{\small (b) Online}
\end{minipage}
\caption{Figure~\ref{fig:cost} with the cost counted in tokens: those of the steered continuation and the mechanism call, minus, online, those of the replaced remainder.}
\label{fig:tokens}
\end{figure}

\begin{table}[!htbp]
\centering\small\setlength{\tabcolsep}{4.5pt}
\caption{Cost of the uncertainty signals per agent step (AppWorld): tokens read or written in addition to the agent's own. Re-read: the scoring pass reads the agent's context again; cached: the context is kept in a prefix cache and only the appended question or prompt is read. Probabilities and entropies come from the pass that generated the action.}
\label{tab:uq_cost}
\begin{tabular}{llrrrr}\toprule
& & \multicolumn{2}{c}{Gemma} & \multicolumn{2}{c}{Qwen}\\\cmidrule(lr){3-4}\cmidrule(lr){5-6}
& Extra pass & re-read & cached & re-read & cached\\\midrule
Probability, entropy & none & 0 & 0 & 0 & 0\\
PMI & action only & 150 & 150 & 118 & 118\\
$P(\mathrm{True})$ & context + action + question & 7{,}559 & 33 & 8{,}857 & 37\\
Black-box & context + action + prompt & 7{,}604 & 75 & 8{,}903 & 75\\
\textcolor{gray}{Agent's own step} & & \multicolumn{2}{c}{\textcolor{gray}{7{,}529}} & \multicolumn{2}{c}{\textcolor{gray}{8{,}828}}\\\bottomrule
\end{tabular}
\end{table}

\begin{figure}[!htbp]
\centering
\begin{minipage}[b]{0.35\linewidth}\centering
\includegraphics[width=\linewidth]{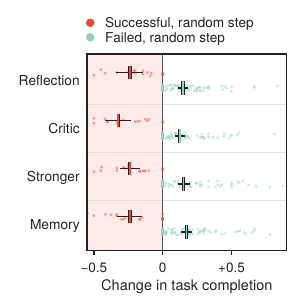}\\[1pt]{\small (a)}
\end{minipage}\hfill
\begin{minipage}[b]{0.222\linewidth}\centering
\includegraphics[width=\linewidth]{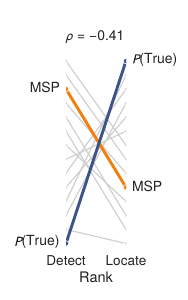}\\[1pt]{\small (b)}
\end{minipage}\hfill
\begin{minipage}[b]{0.415\linewidth}\centering
\includegraphics[width=\linewidth]{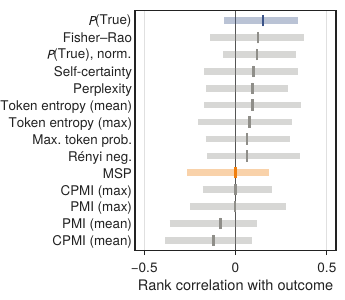}\\[1pt]{\small (c)}
\end{minipage}
\caption{Figure~\ref{fig:s3} for Qwen.}
\label{fig:s3_qwen}
\end{figure}

\begin{figure}[!htbp]
\centering
\includegraphics[width=0.91\linewidth]{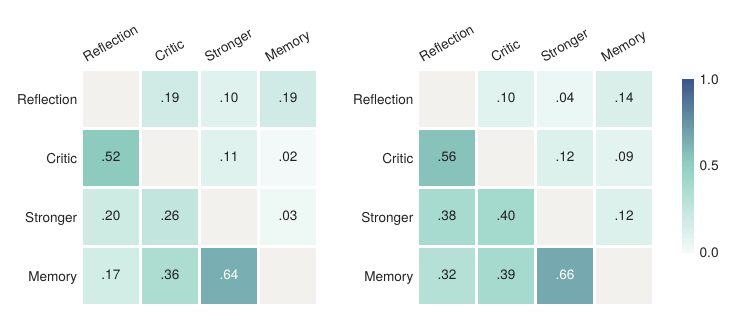}
\caption{Agreement between mechanisms on failed trajectories (left: Gemma; right: Qwen). Below the diagonal: Spearman correlation of best outcomes across trajectories; above: across steps, averaged over trajectories.}
\label{fig:mechanism_matrix}
\end{figure}

\end{document}